\documentclass[eat,twocolumn]{jmlr}

\usepackage{float}

\usepackage{longtable}
\usepackage[margin=1.1in]{geometry}
\usepackage{booktabs}
\usepackage[load-configurations=version-1]{siunitx} 

\theorembodyfont{\upshape}
\theoremheaderfont{\scshape}
\theorempostheader{:}
\theoremsep{\newline}

\usepackage{float}
\jmlrvolume{193}
\firstpageno{1}

\jmlryear{2022}
\jmlrworkshop{Machine Learning for Health (ML4H) 2022}

\title[Extended Survival NAM]{Extending the Neural Additive Model for Survival Analysis with EHR Data
}

  \author{\Name{Matthew Peroni\nametag{\thanks{Majority of work completed while working at Enolink.}}}
  \Email{mperoni1@mit.edu}\\
  \addr Operations Research Center, Massachusetts Institute of Technology\\
  \addr Enolink Inc.\\
  \Name{Marharyta Kurban}
  \Email{kurbanrita@gmail.com}\\
  \addr Enolink Inc.\\
  \Name{Sun Young Yang} \Email{syyang@snuh.org}\\
  \Name{Young Sun Kim} \Email{youngsun@snuh.org}\\
  \Name{Hae Yeon Kang} \Email{65476@snuh.org}\\
  \Name{Ji Hyun Song} \Email{65428@snuh.org}\\
  \addr Department of Internal Medicine, Seoul National University Hospital Healthcare System Gangnam Center
  }

\begin{document}

\maketitle

\begin{abstract}
With increasing interest in applying machine learning to develop healthcare solutions, there is a desire to create interpretable deep learning models for survival analysis. In this paper, we extend the Neural Additive Model (NAM) \cite{nam} by incorporating pairwise feature interaction networks and equip these models with loss functions that fit both proportional and non-proportional extensions of the Cox model. We show that within this extended framework, we can construct non-proportional hazard models, which we call TimeNAM, that significantly improve performance over the standard NAM model architecture on benchmark survival datasets. We apply these model architectures to data from the Electronic Health Record (EHR) database of Seoul National University Hospital Gangnam Center (SNUHGC) to build an interpretable neural network survival model for gastric cancer prediction. We demonstrate that on both benchmark survival analysis datasets, as well as on our gastric cancer dataset, our model architectures yield performance that matches, or surpasses, the current state-of-the-art black-box methods.
\end{abstract}
\begin{keywords}
Interpretable ML, Survival Analysis, Neural Additive Model
\end{keywords}

\section{Introduction}
\label{sec:intro}

Motivated by a series of well-documented failures of machine learning (ML) models being deployed in real-world systems with harmful biases and costly inaccurate predictions \cite{bad-ml-1, bad-ml-2, bad-ml-3}, and the discovery that post-hoc black-box model explainers are not always truthful to the underlying model \cite{shap-problems},  there has been significant interest in developing interpretable, white-box machine learning algorithms \cite{rudin-stop}, especially for high-stakes decision making, such as healthcare \cite{sharp-thresholds}. While development in this direction has been promising, there still exists a gap in the literature that studies the use of interpretable ML methods, especially deep learning methods, for survival analysis with censored data. Since healthcare datasets are often censored and extremely imbalanced, standard classification and regression models, which is what the majority of recent developments in interpretable ML have been designed for, can be ineffective. A promising approach in this direction is the Neural Additive Model (NAM) introduced in \cite{nam}, which offers a directly intelligible and flexible model that can be easily configured to work with survival data. 

In this paper, we extend NAM by incorporating pairwise feature interaction networks and equip these models with loss functions that fit both proportional and non-proportional extensions of the Cox Proportional Hazards (CoxPH) model \cite{cox}. We show that not only does this extension allow us to model crucial feature interactions, it also allows us to model the risk contribution of a feature over time. We find that adding this flexibility has a significant impact on model accuracy, while also allowing us to discover temporal patterns within the data. A primary goal of this paper is to demonstrate the ability of interpretable models to achieve accuracy comparable to state-of-the-art black-box models on challenging real-world clinical survival data. In this vein, we apply the extended NAM model architecture to data from the Electronic Health Record (EHR) database of Seoul National University Hospital Gangnam Center (SNUHGC) to build an interpretable survival model for gastric cancer prediction. We demonstrate that on our gastric cancer dataset, the extended NAM model yields performance that matches, or surpasses, the current state-of-the-art black-box survival methods, such as DeepSurv \cite{deepsurv} and XGBoost \cite{xgboost}.

\section{Related Work}
\label{sec:rel_work}

This paper builds directly upon the Neural Additive Model architecture given in \cite{nam}. To the authors' knowledge, the only other paper to date that has investigated applying NAM to survival data is \cite{survnam}, which uses a NAM model as a surrogate for a black-box deep learning model that is trained on survival data. Our work differs in that we extend the NAM architecture in order to create interpretable models that can replace black-box models entirely, rather than using them as surrogates. Other interpretable deep learning methods have been developed, such as the Neural Interaction Transparency (NIT) model \cite{nit}. The NIT model takes a top-down approach to modeling feature interactions, letting the model determine what interactions to include during the training process. In contrast, our extended NAM approach, outlined in Section \ref{subsec:models}, takes a bottom-up approach, requiring the feature interactions to be identified and specified manually. While the NIT approach is more efficient for discovering useful feature interactions, we find that when working with healthcare survival data, there is a desire to control the feature interactions being modeled. Further, in all experiments we run, only a few interaction pairs are significant enough to include in the final model. Therefore, a bottom-up approach guided by domain experts seems to suit this domain better.
\section{Methods}
\label{sec:methods}
 We cover the extended NAM architecture and detail how it can be equipped for survival analysis. For a brief overview of survival data and analysis, see Appendix \ref{appendix:survival}.
 

\subsection{Models}
\label{subsec:models}
We extend the NAM model introduced in \cite{nam} by modeling select feature interactions. The extended NAM (NA2M) belongs to a class of models called Generalized Additive Models plus Interactions (GA2M) \cite{ga2m} which take the form $g(E[y]) = \sum f_i(x_i) + \sum f_{ij}(x_i, x_j)$ where $g$ is the link function (e.g. identity, logarithm). The NA2M model architecture learns a linear combination of feature networks $f_i^\theta$ that attend to a single feature $i$ and feature pair networks $f_{ij}^\theta$ that attend a pair of features, parameterized by $\theta$. We denote by $s \subseteq \{(i,j) : 1 \leq i < j \leq n\}$ the set of feature pairs used by a given model. The networks are trained jointly using backpropagation. Details of model training are given in Appendix \ref{appendix:training}. In this work, we fit the NA2M model as the log-relative risk function in the CoxPH model, such that $h_r(\mathbf{x}; \theta) = \sum f_i^\theta(x_i) + \sum_{(i,j) \in s} f_{ij}^\theta(x_i, x_j)$. Allowing the model to capture second-order pair interactions can improve accuracy and enhance clinical investigations by illustrating the predictive patterns present in the data for feature pair interactions.

A major benefit we found of extending to the NA2M architecture for survival analysis is that by treating the event time $T$ as a feature, we can create a non-proportional Cox-Time model as defined in \cite{pycox} by learning a time-varying log-relative risk function as $h_r(\mathbf{x}, t; \theta) = \sum f_i^\theta(x_i, t) + \sum f_{ij}^\theta(x_i, x_j)$. 
We refer to this architecture as TimeNAM, or TimeNA2M if feature pairs are included. That is, while both model formats are based on the NA2M model architecture, TimeNAM is comprised solely of feature-interaction nets where each net considers a feature-time interaction, whereas TimeNA2M permits feature-time interactions and, additionally, feature pair interactions, as in the NA2M model. In both cases, we are modeling the risk contribution of each feature over time. As we will demonstrate in the results (\ref{section:results}), this added expressiveness can lead to a significant boost in accuracy while preserving the interpretability of the model.

\begin{table*}[t!]
\floatconts
  {tab:example-booktabs}
  {\caption{C-Index for models across datasets. Means and standard deviations reported from stratified 5-fold cross validation.}}
  {\resizebox{\textwidth}{2.2cm}{\begin{tabular}{llllll}
  \toprule
  \bfseries Model & \bfseries METABRIC & \bfseries SUPPORT & \bfseries GBSG & \bfseries WHAS & \bfseries SNUHGC\\
  \midrule
  Linear Cox & $63.45 \pm 1.19$ & $57.04 \pm 0.94$ & $66.26 \pm 1.32$ & $74.49 \pm 2.63$ & $76.29 \pm 3.33$ \\\hline
  NAM & $64.11 \pm 1.26$ & $60.50 \pm 1.10$ & $65.76 \pm 2.72$ & $74.87 \pm 2.08$ & $73.4 \pm 5.16$\\
  NA2M & $64.44 \pm 1.24$ & $60.76 \pm 0.78$ & $67.65 \pm 0.96$ & $75.64 \pm 2.72$ & $74.31 \pm 3.80$\\
  TimeNAM & $66.79 \pm 2.30$ & $61.95 \pm 0.84$ & $67.76 \pm 1.37$ & $74.58 \pm 1.82$ & $76.21 \pm 1.95$\\
  TimeNA2M & $66.07 \pm 2.26$  & $62.21 \pm 0.73$ & $68.25 \pm 1.44$ & $75.51 \pm 2.09$ & $76.62 \pm 3.00$\\\hline
  DeepSurv & $64.98 \pm 0.80$ & $60.97 \pm 1.19$ & $67.21 \pm 0.98$ & $80.94 \pm 1.15$ & $76.26 \pm 3.36$\\
  XGBoost & $64.65 \pm 1.34$ & $61.97 \pm 0.98$ & $67.47 \pm 1.81$ & $86.99 \pm 0.98$ & $76.50 \pm 1.67$\\
  \bottomrule
  \end{tabular}}}
\label{table:benchmark}
\end{table*}

\section{Results}
\label{sec:results}
\subsection{Data}

We evaluate our models on four benchmark clinical survival datasets: METABRIC \cite{metabric-1, metabric-2}, SUPPORT \cite{support}, WHAS \cite{whas}, and Rotterdam \& GBSG \cite{gbsg-1, gbsg-2, gbsg-3}. We use the versions of these datasets produced by \cite{deepsurv} without any additional processing. 

We also evaluate our models on comprehensive medical annual check-up data, including endoscopic findings and blood test results, collected from 129K+ patients who visited one of the largest medical screening facilities in South Korea. A cancer diagnosis was determined based on the stomach biopsy results of the samples taken during endoscopy procedures. The final cohort consists of 59,540 negative and 248 positive patients. See \appendixref{apd:second} for more details. We refer to this dataset as SNUHGC.

\subsection{Experiment Results}
\label{section:results}
In the following, we compare our extended NAM models against the linear Cox model \cite{cox}, XGBoost \cite{xgboost}, and DeepSurv \cite{deepsurv} on our benchmark datasets and SNUHGC data. We evaluate the models using the \textbf{concordance index} (c-index) metric, as defined in \cite{cindex}. The c-index is the standard metric for survival analysis and is the equivalent of the ROC AUC score for survival data. Hyperparameter tuning for each model was conducted using Bayesian Optimization \cite{bayes-opt}, details can be found in appendix \ref{appendix:hyperparam}.  To identify important features pairs for inclusion in the extended NAM models, we use the FAST algorithm formulated in \cite{fast} with minor modifications for public datasets and clinical expertise for the SNUHGC dataset. For each dataset, we include at most five feature interactions in the extended NAM models.

\subsubsection{Benchmark Experiments}
Our main results are summarized in Table \ref{table:benchmark}. In two instances, Metabric and GBSG, we see that the TimeNAM models actually outperform our comparison models. This is possible because TimeNAM is able to consider time as a feature, whereas the XGBoost and Deepsurv models, as originally implemented, are based on the Cox proportional hazards model. These other models could be modified to fit the Cox-Time framework, but our primary goal in this work is to demonstrate the flexibility of the extended NAM models and their ability to match the performance of pre-existing black-box models.

We also observe that some datasets, such as GBSG, have significant pairwise interactions such that the NA2M model outperforms the baseline NAM model, while other datasets, such as Metabric and Support, have significant time-varying risk factors such that the TimeNAM models outperform the baseline NAM model. In the case of GBSG, we see that there are predictive patterns in both the pairwise interactions and between features and time, such that both NA2M and TimeNAM improve upon the baseline, and the combination of TimeNAM plus feature interactions yields an even better model. 

The one outlier in these results is the WHAS dataset, where none of the NAM based models perform much better than the linear Cox model or come close to the black-box models. Upon further investigation we discovered that, as noted in \cite{pycox}, the dataset is a case-control dataset, such that there are duplicate records of patients in the data. Therefore, it is possible that without the constraints placed on the NAM models, the DeepSurv and XGBoost models are more capable of memorizing data that is present in both the training and test sets. Indeed, we can see in Table \ref{table:hyperparam-xgboost} in Appendix \ref{appendix:hyperparam} that the hyperparameter search for XGBoost on the WHAS dataset selected the highest possible value for the number of estimators, suggesting that overfitting to the training data (since boosted trees are trained on the residual) was beneficial in this case.

\subsubsection{SNUHGC Experiments}
The last column of Table \ref{table:benchmark} summarizes the model performance on the SNUHGC dataset. The table indicates that a simple NAM model has the lowest performance due to its inability to model feature interactions. Adding feature pairs improved the performance slightly, but the performance was still lower than that of the benchmark models. The addition of the time component along with the feature interactions(TimeNA2M) yielded performance equivalent to DeepSurv and XGBoost.

\begin{figure}[h!]
    \centering
    \includegraphics[width=\linewidth]{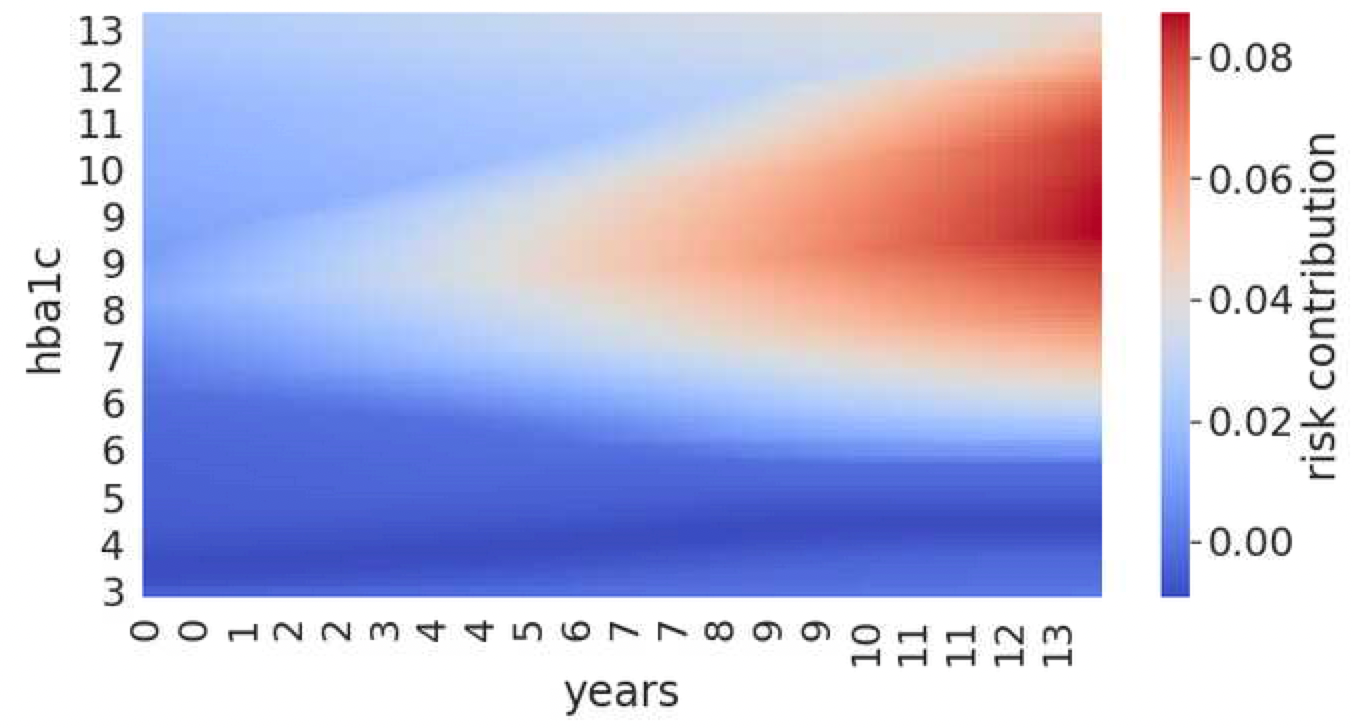}
    \caption{HbA1c risk contribution over time for SNUHGC. The Plot depicts the exact output from the feature net that models the risk contribution of HbA1c over time in the TimeNA2M model fit to this dataset.}
    \label{fig:hba1c}
\end{figure}

Figure \ref{fig:hba1c} provides an illustrative example visualization of the TimeNA2M output. It shows the risk contribution of HbA1c and  indicates that poor glycaemic control may increase the risk of gastric cancer over time, which is in line with the experimental findings \cite{hba1c}.

\begin{figure*}[h!]
    \centering
    \includegraphics[width=\linewidth]{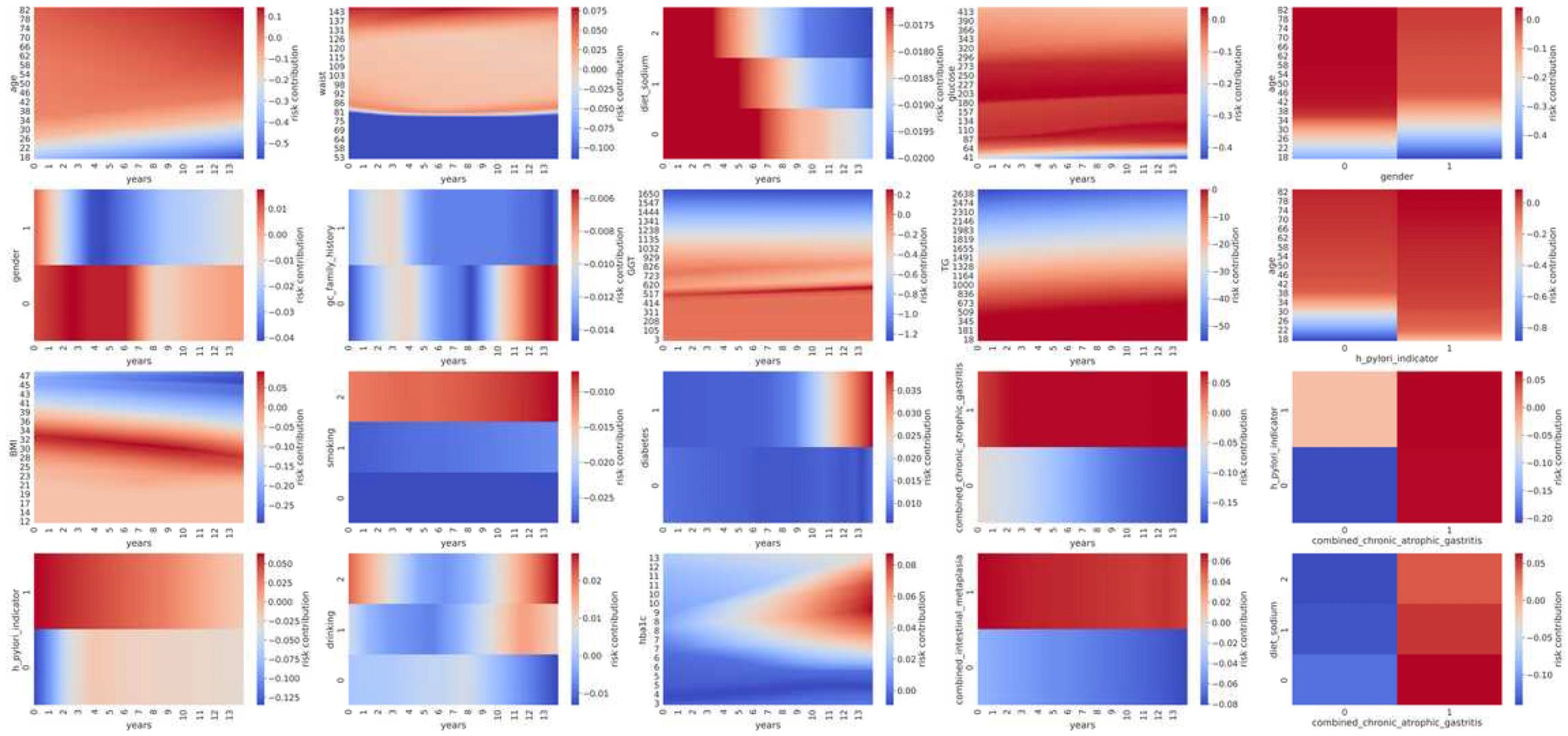}
    \caption{TimeNA2M feature network outputs for SNUHGC. Each plot corresponds to an individual feature or feature pair network.}
    \label{fig:tn_fp}
\end{figure*}

Figure \ref{fig:tn_fp} shows the full output of the TimeNA2M model trained on SNUHGC data. It illustrates the risk contribution of each feature over time and shows feature interaction pairs selected by PIs in the last column: gender - age, age - h. Pylori, h. Pylori - atrophic gastritis, diet sodium - gastritis. It is worth noting that not all plots can be explained easily. For example, the figures show that higher salt intake is associated with protective properties, which contradicts the research \cite{salt}. Such a mismatch could be explained by the fact that patients' sodium intake data were self-reported and, thus, might be unreliable. Another explanation is that patients with gastric conditions tend to adjust their salt intake. Thus, reduced sodium intake could be a proxy for the presence of other conditions. However, the model correctly identifies atrophic gastritis, intestinal metaplasia, and h. Pylori infection as the main contributing factors as illustrated by the feature importance plot (Figure \ref{fig:feat_imp}) and gets the direction of their effect correct. It also correctly identifies most lifestyle factors such as smoking and drinking. From this starting point, however, we can work with clinicians to understand the data and refine the model.  

\begin{figure*}[h!]
    \centering
    \includegraphics[width=\linewidth]{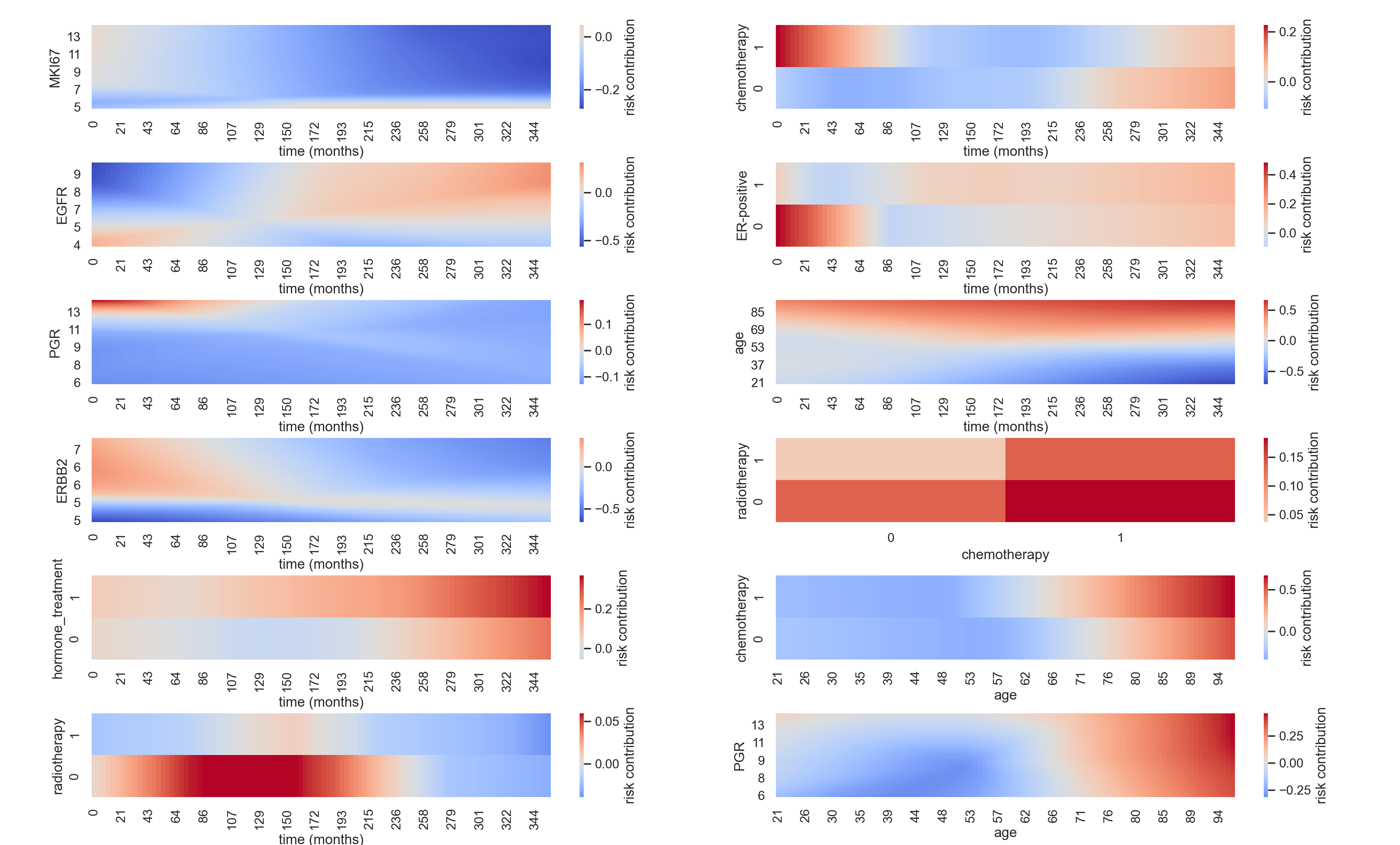}
    \caption{TimeNA2M feature and feature pair network outputs for METABRIC.}
    \label{fig:metabric}
\end{figure*}

As the model is solely reflective of the patterns present in this dataset, which is also highly imbalanced, we observe some of the feature (pair) networks produce risk functions that could not be explained by domain expertise, discussed further in \appendixref{apd:second}. However, since these NAM models can be easily modified with the guidance of domain knowledge and the plots (such as \ref{fig:hba1c}) are an \textit{exact} representation of the model, we believe these models provide an advantage over the black-box models. 



\subsubsection{METABRIC Case Study}
To further illustrate the interpretability of these extended NAM models, we provide the TimeNA2M model visualization for METABRIC in Figure \ref{fig:metabric}. The METABRIC dataset consists of nine features total, including age, four gene and protein expression features (MKI67, EGFR, PGR, ERBB2), and four relavant clinical indicator variables (hormone treatment, radiotherapy, chemotherapy, and ER positive) for patients with breast cancer. There are 1,980 patients in the dataset, of which 57.72 percent died during the study. As discussed in section \ref{section:results}, there are predictive patterns in the relationship between features and time that boost the performance of the TimeNAM models. We can see this reflected in the feature function plots, as the relative risk contribution of EGFR, ERBB2, radiotherapy, chemotherapy, and being ER-Positive all clearly vary over time. The included feature pairs do not significantly contribute to the performance of the model, but we do observe some clear interaction between radiotherapy and chemotherapy. In this case, the modeled interaction tells us that to the model, a patient undergoing chemotherapy treatments without radiotherapy indicates the patient is at higher risk of death than if they were treated with only radiotherapy or with both treatments. While the model has no notion of causality, this interpretability can be very useful for collaborations with clinicians with domain knowledge. As mentioned in \ref{section:results}, the plots in Figure \ref{fig:metabric} exactly represent the model, so the model itself can be easily modified by essentially modifying theses plots.
\section{Discussion and Future Work}
\label{sec:discussion}
In this work, we demonstrated that interpretable, white-box models are capable of achieving performance comparable to black-box models for survival analysis. The modularity, flexibility, and ability to make precise modifications to these extended NAM models enable meaningful collaboration with domain experts where all parties can understand the model and suggest pinpoint modifications. Going forward we plan to continue our work by: (1) Comparing the extended NAM model visualizations to SHAP explanations on the same clinical datasets (we begin to explore this in Appendix \ref{appendix:shap-snuh} and \ref{appendix:metabric-shap}). (2) Exploring other modifications to NAM, such as the recently published Neural Basis Model \cite{NBM}. (3) Evaluating these models on additional datasets.



\appendix

\section{A Brief Summary of Relevant Survival Analysis}
\label{appendix:survival}
\subsection{Survival Data}
We consider survival data of the form $(\mathbf{x}, T, E)$ where $\mathbf{x} = (x_1, x_2, \dots, x_K)$ is the input covariates with $K$ features, $T$ is a failure event time, and $E$ is an event indicator. When $E = 1$, this indicates that an event (e.g. death of patient, development of disease, etc.) has occurred at time $T$ from the beginning of the study. When $E = 0$, this indicates that no event has occurred by the last recorded time $T$, which can be the end of the study, or the last contact with the patient. The latter is considered \textit{right-censored data}, and is generally considered missing data when applying standard regression or classification techniques. Given the prevalence of censored data in real-world clincial datasets, ignoring this data can significantly bias the model. Therefore, we want to use methods that incorporate this censored data, which forms the basis of survival analysis.

\subsection{Survival Analysis}
We give a brief description of the essential components of survival analysis for our work, see \cite{survival-overview} for a complete overview. The two essential components of survival analysis are the survival function, $S(t) = Pr(T > t)$, representing the probability that an individual will 'survive' beyond time $t$, and the hazard function $h(t)$, given by
$$h(t) = \lim_{\delta \to 0} \frac{Pr(t \leq T < t + \delta | T \geq t)}{\delta}$$
which represents the probability an individual will not 'survive' beyond time $T$, given that they have already 'survived' up until that time. It is known that we can relate the cumulative hazard function, $H(t) = \int_0^t h(s) ds$ to the survival function $S(t)$ by the equation $S(t) = \exp{[-H(t)]}$.
We base our work on two models of hazard function: (1) the Cox Proportional Hazards Model (CoxPH) \cite{cox}, which represents the hazard function as
$$h(t | \mathbf{x}) = h_0(t)\exp{[h_r(\mathbf{x})]}$$
where $h_0(t)$ is a non-parametric baseline hazard function, and $\exp{[h_r(\mathbf{x})]}$ is the \textit{relative risk function}. (2) The non-proportional Cox Time Model (Cox-Time) \cite{pycox} which represents the hazard function as 
$$h(t | \mathbf{x}) = h_0(t)\exp{[h_r(\mathbf{x}, t)]}$$
where the relative risk function can depend on time.

\section{Supplementary Materials for Model Evaluation}\label{apd:first}

\subsection{Model Training}
\label{appendix:training}
We train our models on datasets of the form $\{\mathbf{x}^{(i)}, E^{(i)}, T^{(i)}\}_{i=1}^N$ with $N$ individuals. Suppose we have $M$ total feature and feature-pair networks in the model. We train the model using an extension of the NAM loss function derived in \cite{nam} given by
$$\mathcal{L}(\theta) = \mathbb{E}_{\mathbf{x}, t, e \sim D} [l(\mathbf{x}, t, e ; \theta) + \lambda_1 \eta(\mathbf{x};\theta)] + \lambda_2 \gamma(\theta)$$
where
\begin{align*}
    \begin{split}
        \eta(\mathbf{x};\theta) = &\\
        \frac{1}{M} \sum_x &(\sum_k(f_k^\theta(x_k))^2 + \sum_{(i,j) \in s}(f_{ij}^\theta(x_i,x_j))^2)
    \end{split}
\end{align*}
is the output regularization and $\gamma(\theta)$ is the weight decay. In our work, we let $l(\cdot; \theta)$ be either the loss function derived in \cite{pycox} for the CoxPH model
\begin{multline*}
l(\cdot; \theta) = \\ 
\frac{1}{n} \sum_{i : E^{(i)} = 1}\log(\sum_{j \in \tilde R^{(i)}} \exp[h_r(\mathbf{x}^{(i)}; \theta) - h_r(\mathbf{x}^{(j)}; \theta)])
\end{multline*}
where $h_r(\mathbf{x}; \theta) = \sum f_i^\theta(x_i) + \sum f_{ij}^\theta(x_i, x_j)$ and $n$ is the number of individuals with a recorded event, or the loss function for the Cox-Time model
\begin{multline*}
l(\cdot; \theta) =\\
\frac{1}{n} \sum_{i : E^{(i)} = 1}\log(\sum_{j \in \tilde R^{(i)}} \exp[h_r(\mathbf{x}^{(i)}, T^{(i)}; \theta)\\ - h_r(\mathbf{x}^{(j)}, T^{(j)}; \theta)])
\end{multline*}
where $h_r(t, \mathbf{x}; \theta) = \sum f_i^\theta(x_i, t) + \sum f_{ij}^\theta(x_i, x_j)$.

It is also worth briefly discussing the relative training time of this model compared to our benchmark models. The standard linear Cox proportional hazards model is the most efficient model, and can solve with n = O(1000) records and p = O(10) features in O(1s). Similarly, XGBoost is very efficient, and can converge in O(1s) for the public datasets considered in this work, and scales well as n and p increase. The neural network models considered do take longer to train, taking between O(1s) to O(1min) to train on the public datasets we used, depending on the hyperparameters and size of the problem. It is also important to note that for the non-proportional time models (TimeNAM and TimeNA2M), predictions are more computationally expensive than in the proportional case. In our experience however, the prediction run-time for TimeNAM is still fast (O(10ms)). For most applications in health care, training time on the order of minutes is still acceptable, and we also note that the NAM-based models are easily modifiable after training, whereas other models such as XGBoost would require making modifications to the training data or hyperparameters and retraining the model.
\subsection{Hyperparameters}
\label{appendix:hyperparam}
\noindent \textbf{Extended NAM.} We trained all the extended NAM models with a batch size of 256. We identify one set of parameters for each dataset, and use these parameters across variations of the extended NAM architecture. We tune the dropout in the set $\{0, 0.05, 0.1, 0.2, 0.3, 0.4, 0.5\}$, feature dropout in the set $\{0, 0.05, 0.1, 0.2, 0.3\}$, weight decay in the interval $[1\mathrm{e}{-6}, 1\mathrm{e}{-4}]$, output regularization in the interval $[0.001, 0.1]$, and learning rate in the interval $[0.001, 0.1]$. We tune the hidden sizes (for each feature or feature pair network) in the set $\{[16], [64], [128], [256], [512], [64,32], [32,16,8]\}$ and activation functions in the set $\{\text{relu}, \text{exu}, \text{sigmoid}\}$. Table \ref{table:hyperparam-nam} gives the hyperparameters used in our model evaluation.

\begin{table*}[t]
\floatconts
  {tab:example-hline}
  {\caption{Parameters used for extended NAM models}}%
  {%
    \begin{tabular}{| l | l | l |l|l|l|}
    \hline
    \abovestrut{2.2ex}\bfseries Hyperparameter/Dataset & \bfseries WHAS & \bfseries SUPPORT & \bfseries Metabric & \bfseries GBSG & \bfseries SNUHGC \\\hline
    \abovestrut{2.2ex}Dropout & 0 & 0 & 0.05 & 0 & 0.2\\\hline
    Feature Dropout & 0 & 0.3 & 0 & 0 & 0\\\hline
    Weight Decay ($\lambda_2$)& 1e-6 & 3.82e-6 & 1e-6 & 1e-4 & 1e-6\\\hline
    Output Reg. ($\lambda_1$) & 0.001 & 0.0466 & 0.001 & 0.001 & 0.0297\\\hline
    Activation & relu & relu & relu & relu & relu\\\hline
    Hidden Size & 512 & 16 & 64,32 & 128 & 32,32\\\hline
    \belowstrut{0.2ex}Learning Rate & 0.001 & 0.001 & 0.001 & 0.001 & 0.022\\\hline
    \end{tabular}
  }
\label{table:hyperparam-nam}
\end{table*}

\noindent \textbf{DeepSurv.} We use the open-source implementation of DeepSurv and use the parameter values identified in \cite{deepsurv} for the benchmark datasets.

\noindent \textbf{XGBoost} We use XGBoost with the objective parameter "survival:cox" that fits the model such that the output is the log-relative risk function $h_r$. We tune the \textit{eta} parameter in the set $\{0.001, 0.005, 0.01\}$, the \textit{max depth} parameter in the set $\{3, 4, 5\}$, the \textit{subsample} parameter in the set $\{0.5, 0.7, 0.9\}$, the \textit{num estimators} parameter in the set $\{100, 200, 300, 400, 500\}$, the \textit{min child weight} parameter in the set $\{1, 5, 10\}$, the \textit{gamma} parameter in the set $\{0, 0.5, 1\}$, the \textit{colsample bytree} parameter in the set $\{0.5, 0.7, 0.9\}$, and the \textit{alpha} and \textit{lambda} parameters in the set $\{0, 0.5, 1\}$. The \textit{tree method} is set to either 'auto' or 'hist'. Table \ref{table:hyperparam-xgboost} gives the hyperparameters used in our model evaluation.

\noindent \textbf{Linear Cox} We use the open source implementation of the Cox linear regression model provided by the lifelines python package \cite{lifelines} without modification. 

\begin{table*}[htbp]
\floatconts
  {tab:example-hline}
  {\caption{Parameters used for XGBoost models}}%
  {%
    \begin{tabular}{| l | l | l |l|l|l|}
    \hline
    \abovestrut{2.2ex}\bfseries Hyperparameter/Dataset & \bfseries WHAS & \bfseries SUPPORT & \bfseries Metabric & \bfseries GBSG & \bfseries SNUHGC\\\hline
    \abovestrut{2.2ex}Eta & 0.01 & 0.05 & 0.01 & 0.001 & 0.01\\\hline
    Max Depth & 5 & 5 & 4 & 5 & 5\\\hline
    Subsample & 0.9 & 0.7 & 0.7 & 0.9 & 0.5\\\hline
    Num. Estimators & 500 & 200 & 300 & 200 & 200\\\hline
    Min Child Weight & 1 & 10 & 10 & 10 & 10\\\hline
    Gamma & 0 & 0 & 1 & 0 & 0\\\hline
    Col. Sample By Tree & 0.9 & 0.5 & 0.7 & 0.5 & 0.5\\\hline
    Alpha & 0 & 1 & 0 & 0 & 0\\\hline
    Lambda & 0 & 0.5 & 0 & 0 & 1\\\hline
    \belowstrut{0.2ex}Tree Method & auto & hist & hist & auto & auto\\\hline
    \end{tabular}
  }
\label{table:hyperparam-xgboost}
\end{table*}

\section{Gastric Cancer Case Study}\label{apd:second}

Our data was extracted from the Electronic Health Record (EHR) database of Seoul National University Hospital Gangnam Center (SNUHGC), which is one of the largest medical check-up centers in South Korea. Patients in our study cohort were selected from 129,223 predominantly healthy individuals who visited the center for a regular annual check-up between 2007 and 2020. Patients underwent various clinical tests such as physical examination, blood tests, gastroscopy, and, in some cases, biopsy. All patients completed a self-administered structured questionnaire about their smoking history, alcohol intake, eating habits, family history, and past medical history. A cancer diagnosis was determined based on the stomach biopsy results of the samples taken during endoscopy procedures. 

Our cohort selection process includes several steps. First, we filtered out patients who had fewer than two hospital visits. Then, we excluded patients with diagnoses determined by PIs: total gastrectomy, squamous cell carcinoma, neuroendocrine tumor, GIST, and lymphoma/MALToma. Lastly, we excluded patients outside the age range of interest (18-90 years.) The final cohort consists of 59,540 negative and 248 positive patients.

An initial set of 60+ features was refined to 16 features through a feature selection process informed by clinical expertise and ML algorithms. Among others, our final feature set includes h.pylori infection, chronic atrophic gastritis, and intestinal metaplasia, which are well-known factors contributing to gastric cancer development.

\subsection{SNUHGC SHAP Experiments}
\label{appendix:shap-snuh}
Many post-hoc black-box model explanation tools have been developed recently, and SHAP (SHapley Additive exPlanation) is one of the most common approaches \cite{shap}. In this section, we use the SHAP library to approximate SHAP values for the interpretable TimeNA2M model we trained on SNUHGC data and produce visualizations. Since we know the 'ground-truth' for how this model makes predictions, we can assess the quality of the SHAP feature attributions.

SHAP explains the prediction for any instance as a sum of contributions from its features. The approach was primarily inspired by game theory. The Shapley value of a specific feature is its contribution to the model prediction, weighted and summed over all possible feature value combinations:
\begin{multline*}
    \phi_{i} = \\
        \sum_{S\subseteq{F}\backslash \{i\}} \frac{|S|!(|F|-|S|-1)!)}{|F|!}[f_{S\cup\{i\}}(x_{S\cup\{i\}})\\
        -f_{S}(x_S)]
\end{multline*}

where S represents all possible subsets of the full feature set F that do not contain feature $i$.

As the formula indicates, the exact calculation of SHAP values is computationally challenging. Lundberg and Lee (2017) \cite{shap} describe two approximation approaches, Shapley sampling values and Kernel SHAP, which are based on the LIME approach. 

\begin{figure*}[]
    \centering
    \includegraphics[width=0.75\linewidth]{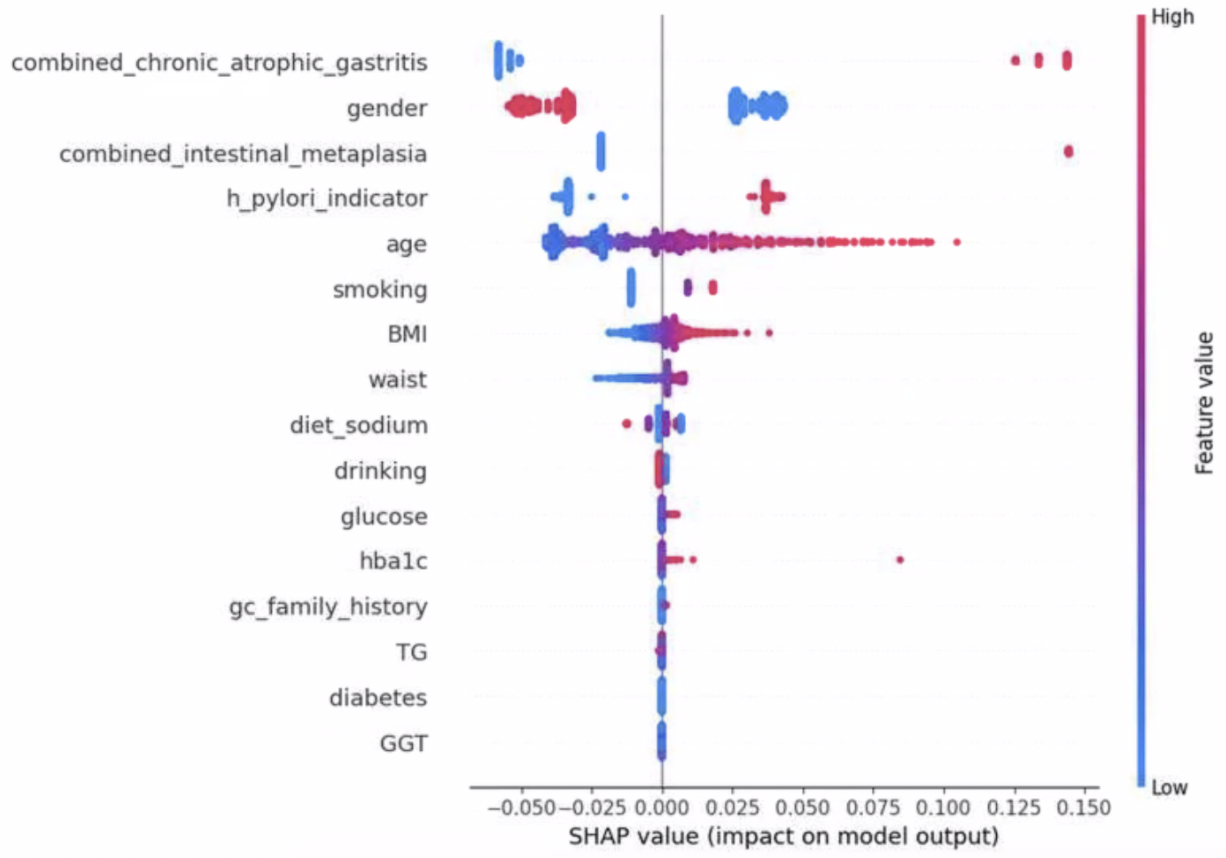}
    \caption{SHAP summary plot for TimeNA2M on SNUHGC dataset.}
    \label{fig:shap-summary-snuh}
\end{figure*}

SHAP summary plot combines feature importance with feature effects. Each point on the summary plot is a Shapley value for a feature and an instance. The position on the y-axis is determined by the feature and on the x-axis by the corresponding Shapley value. The resulting plot for the TimeNA2M model created for the SNUHGC data is given in Figure \ref{fig:shap-summary-snuh}.

The plot indicates that 4 out of 5 top features are correctly identified by SHAP. The direction of the effect is also the same for most features. However, by introducing the time component, we can better stratify and analyze the feature effect, such as in the example of HbA1c. 

\begin{figure*}[]
    \centering
    \includegraphics[width=0.75\linewidth]{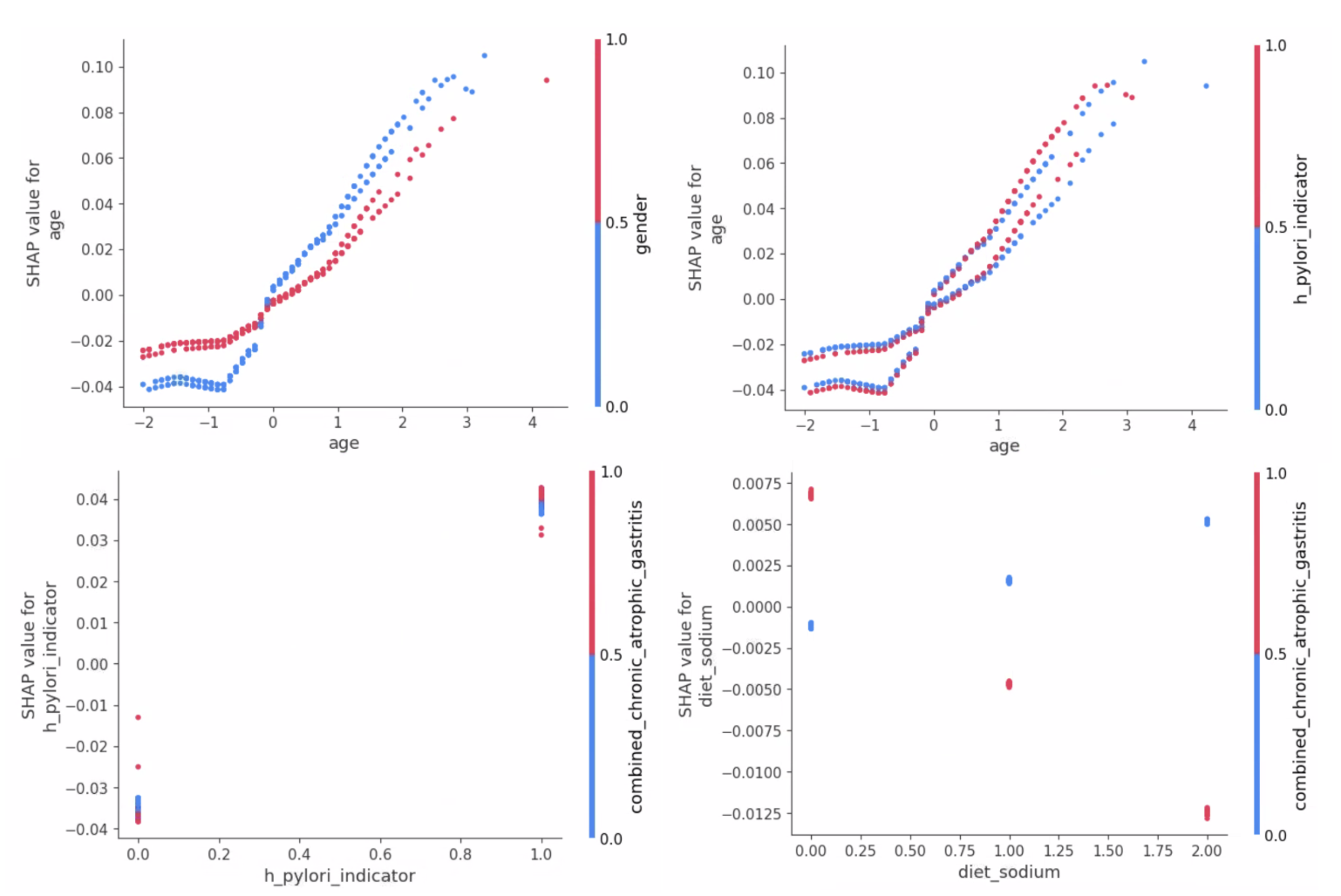}
    \caption{SHAP dependence plots for TimeNA2M on SNUHGC dataset.}
    \label{fig:shap-dependence-snuh}
\end{figure*}

The dependence plots, shown in Figure \ref{fig:shap-dependence-snuh}, require further interpretation. For example, the effect of the gender  seems to flip around the mean age, which is not observed in the ground truth of the underlying model. This example demonstrates that interpretability and post hoc explainability can diverge, highlighting the importance of white box methods.

\section{METABRIC Case Study SHAP Experiments}
\label{appendix:metabric-shap}
\begin{figure*}[h!]
    \centering
    \includegraphics[width=\linewidth]{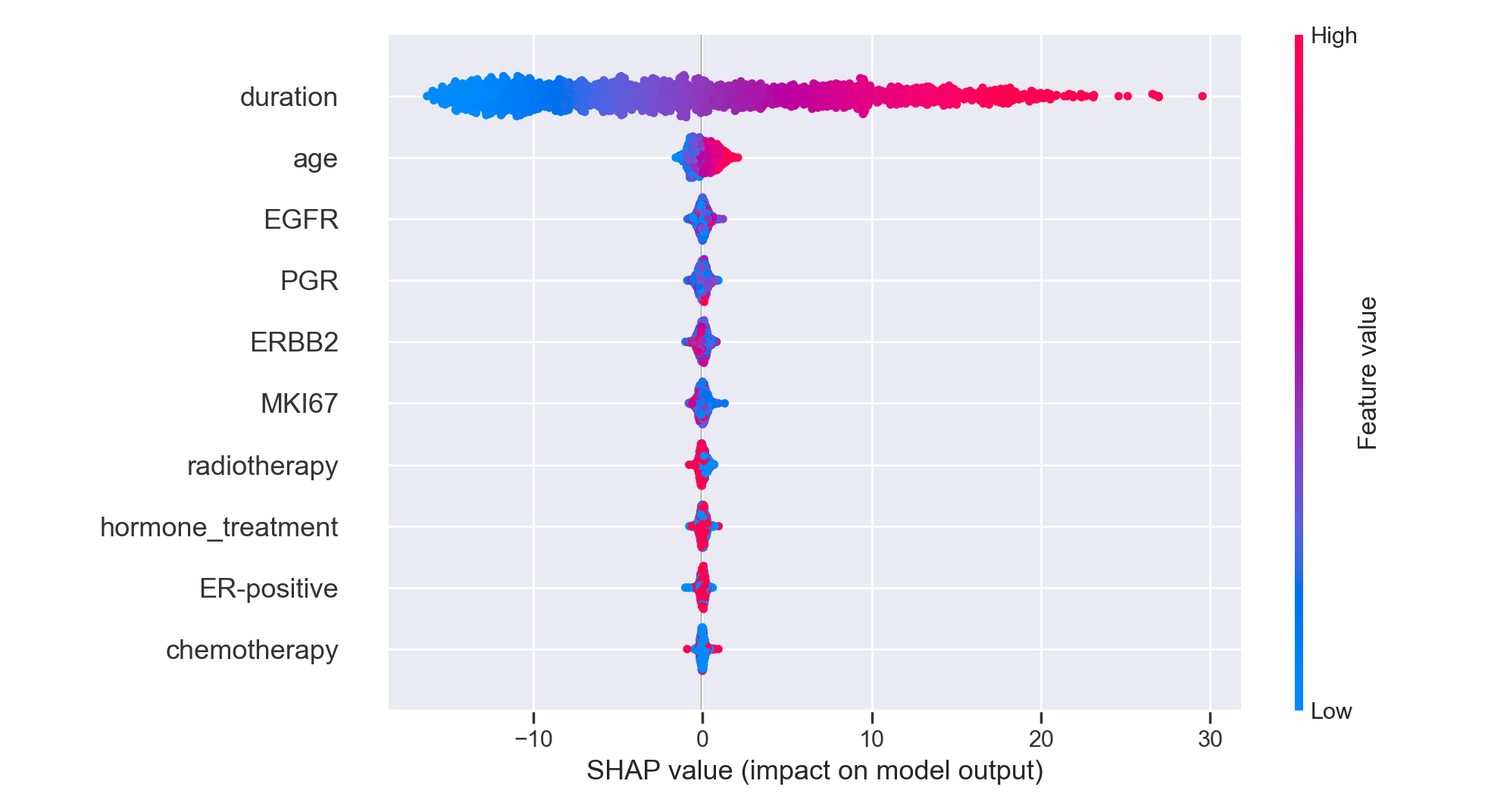}
    \caption{SHAP Summary for TimeNA2M Model on METABRIC dataset.}
    \label{fig:metabric-shap}
\end{figure*}

We tried using SHAP to generate post-hoc explanations of the TimeNA2M model trained on the METABRIC dataset. We used the Permutation SHAP explainer, which is a model agnostic explainer that approximates the Shapley values by iterating over random permutations of the given data. This method was used for efficiency, as the exact SHAP explainer took unreasonably long to fit. The summary plot for the resulting explainer can be seen in Figure \ref{fig:metabric-shap}. We can observe that the explainer identifies duration as the most significant feature by an order of magnitude. While the explainer accurately captures the effect that as duration increases, a patient's risk increases, this is a fairly uninformative result, as it matches general intuition. Further, it seems to mask the contributions of other features. If one did not have an intuition regarding the meaning of the given features, or how the features ought to impact predictions, it is easy to imagine a scenario where one would conclude that the duration feature is the only significant feature in determining risk. The explainer also seems to indicate that, for example, chemotherapy does not have much impact on the risk prediction under this model. In Figure \ref{fig:metabric}, where we see the ground truth regarding how the model uses the chemotherapy feature, we can see a more complicated story, where the model identifies a patient receiving chemotherapy soon after their diagnosis as increasing their risk, while it seems to decrease their risk in later stages. This illustrates how post-hoc explainers can obscure the actual patterns learned by a model.

\begin{figure*}[h!]
    \centering
    \includegraphics[width=\linewidth]{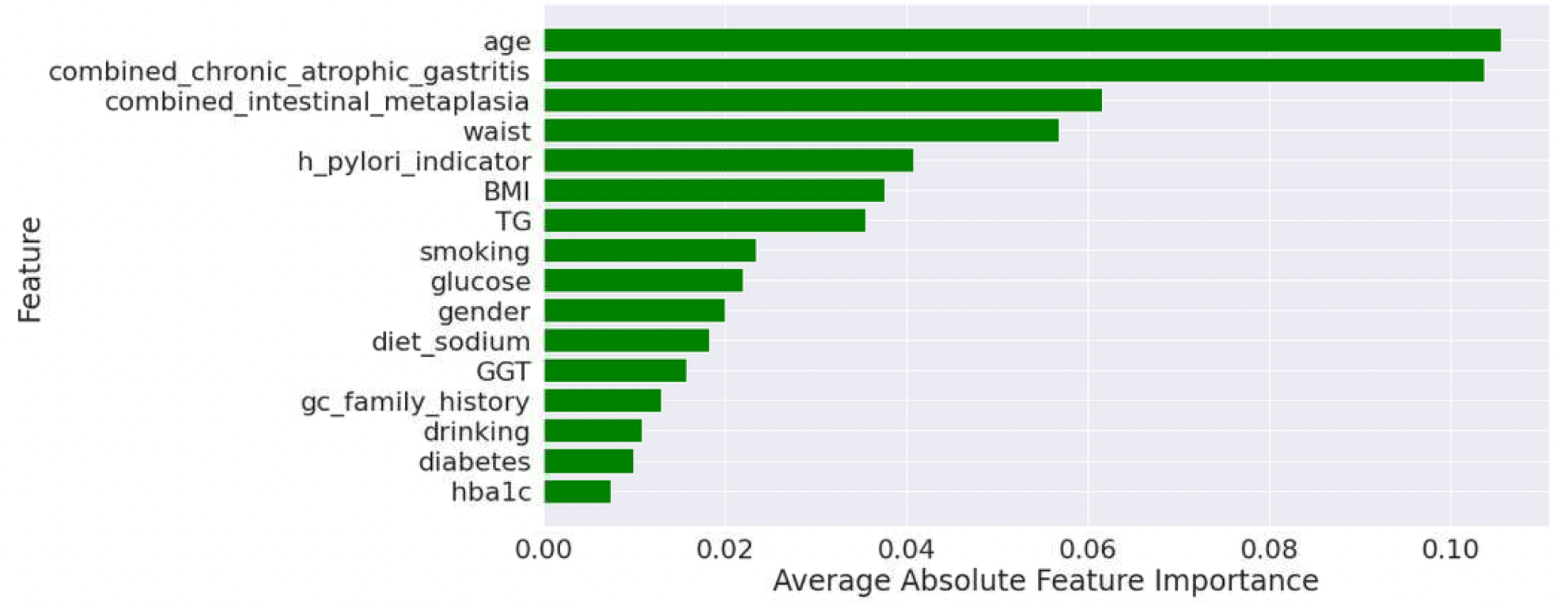}
    \caption{TimeNA2M feature importance for SNUHGC.}
    \label{fig:feat_imp}
\end{figure*}

\end{document}